\documentclass{article}
\usepackage{arxiv}

\usepackage[utf8]{inputenc}
\usepackage[T1]{fontenc}
\usepackage{microtype}
\usepackage{graphicx}
\graphicspath{{./images/}}
\usepackage{amsmath,amssymb}
\usepackage{subcaption}
\usepackage{booktabs}
\usepackage{natbib}
\setcitestyle{authoryear,round}

\usepackage{hyperref}
\hypersetup{hidelinks} % or: colorlinks=true, linkcolor=black, citecolor=black, urlcolor=blue

\usepackage{caption}
\title{Spectral Predictability as a Fast Reliability Indicator for Time Series Forecasting Model Selection}

\author{
  Oliver Wang\\
  Electrical and Computer Engineering\\
  University of California, Los Angeles\\
  owang22@g.ucla.edu
  \And
  Pengrui Quan\\
  Electrical and Computer Engineering\\
  University of California, Los Angeles\\
  prquan@g.ucla.edu
  \And
  Kang Yang\\
  Electrical and Computer Engineering\\
  University of California, Los Angeles\\
  kyang73@g.ucla.edu
  \And
  Mani Srivastava\footnotemark[1]\thanks{Mani Srivastava holds concurrent appointments as a Professor of ECE and CS (joint) at the University of California, Los Angeles, and as an Amazon Scholar at Amazon. This paper describes work performed at UCLA and is not associated with Amazon.}\\
  Electrical and Computer Engineering\\
  University of California, Los Angeles\\
  mbs@ucla.edu
}

\begin{document}
\maketitle

\begin{abstract}
Practitioners deploying time series forecasting models face a dilemma: exhaustively validating dozens of models is computationally prohibitive, yet choosing the wrong model risks poor performance. We show that \emph{spectral predictability}~$\Omega$---a simple signal processing metric---systematically stratifies model family performance, enabling fast model selection. We conduct controlled experiments in four different domains, then further expand our analysis to 51 models and 28 datasets from the GIFT-Eval benchmark. We find that large time series foundation models (TSFMs) systematically outperform lightweight task-trained baselines when $\Omega$ is high, while their advantage vanishes as $\Omega$ drops. Computing $\Omega$ takes seconds per dataset, enabling practitioners to quickly assess whether their data suits TSFM approaches or whether simpler, cheaper models suffice. We demonstrate that $\Omega$ stratifies model performance predictably, offering a practical first-pass filter that reduces validation costs while highlighting the need for models that excel on genuinely difficult (low-$\Omega$) problems rather than merely optimizing easy ones.
\end{abstract}

% -------------------------------------------------------------------------
\section{Introduction}

Large time series foundation models (TSFMs) for time series forecasting promise broad improvements by leveraging massive pretraining \citep{ye2024surveytimeseriesfoundation,li2025tsfm,Liang_2024,ansari2024chronoslearninglanguagetime,gruver2024largelanguagemodelszeroshot}. 
Yet empirical evidence remains mixed; simple baselines such as DLinear often match or surpass complex architectures~\citep{tan2024languagemodelsactuallyuseful,zeng2022transformerseffectivetimeseries,li2025tsfm}. Practitioners face a practical challenge: \emph{how to choose which model to deploy without exhaustively validating every option?}

Comprehensive validation is impractical. Consider a practitioner with a dozen or more candidate models and a new dataset: training and validating all models requires substantial compute, time, and engineering effort. Worse, this process provides no insight into \emph{why} certain models work better, making it difficult to generalize lessons to future datasets.

We propose \textit{spectral predictability}~$\Omega$---a simple, fast-to-compute signal property---as a reliability indicator that narrows the model search space \emph{before} expensive validation begins. Grounded in signal processing, $\Omega$ quantifies the concentration of a series' power spectrum: high~$\Omega$ reflects structured, repeatable patterns; low~$\Omega$ indicates diffuse, irregular signals. Computing $\Omega$ takes seconds on a commodity device, yet we show it systematically stratifies model performance.

\textbf{Our key finding.}
Large zero-shot\footnote{We use GIFT-Eval's model taxonomy where ``zero-shot'' refers to TSFMs deployed with their original pretrained weights. See Large-Scale Analysis Results for full definitions.} models, applied without fine-tuning, show consistent advantages in high-$\Omega$ regimes across diverse domains. 
Practitioners can compute~$\Omega$ to determine whether zero-shot or lightweight models are likely to perform best, reducing validation cost. As $\Omega$ decreases, model performance converges, underscoring the need for methods that better handle difficult (low-$\Omega$) data.

In summary, this paper makes the following contributions:

\begin{itemize}

\item We introduce spectral predictability~$\Omega$ as a fast and interpretable indicator of time-series forecastability, derived from frequency-domain concentration and computable without model training.

\item Controlled experiments on synthetic and real-world datasets (CarbonCast, PEMS, Fitbit) show that forecasting error decreases monotonically with increasing~$\Omega$, confirming that $\Omega$ reflects intrinsic difficulty.

\item Large-scale analysis of 51 models and 28 GIFT-Eval datasets shows that zero-shot TSFMs outperform statistical and deep-learning baselines by up to~60\% in high-$\Omega$ regimes, while the advantage vanishes in low-$\Omega$ settings.

\item We identify the low-$\Omega$ regime as a critical open frontier where all model families struggle, motivating the design of models robust to irregular or weakly periodic signals.
\item We provide actionable guidance for practitioners, showing that computing~$\Omega$ takes seconds yet reliably narrows the model search space, reducing validation cost and improving deployment efficiency.
\end{itemize}

\section{Related Work}

\textbf{Simplicity versus Capacity.} Despite scaling trends \citep{shi2024scalinglawtimeseries,shi2025timemoebillionscaletimeseries}, lightweight baselines remain competitive \citep{zeng2022transformerseffectivetimeseries,miller2024surveydeeplearningfoundation}. Comparative studies \citep{goswami2024momentfamilyopentimeseries,jin2024timellmtimeseriesforecasting} rarely explain \emph{why} performance varies across domains, leaving practitioners without guidance for model selection.

\textbf{LLMs for Time Series.} Methods include direct tokenization, architectural adaptation, and adapter-based fine-tuning \citep{gruver2024largelanguagemodelszeroshot,ansari2024chronoslearninglanguagetime}. Ablations question how much LLM pretraining contributes \citep{tan2024languagemodelsactuallyuseful,jin2024timellmtimeseriesforecasting,elsayed2021reallyneeddeeplearning}. We build on \citet{jin2024timellmtimeseriesforecasting} for our codebase and initial experiments are based on variations on their LLAMA-7B backbone structure, which will be explained further in the Controlled Experiment Results.

\textbf{Forecastability and Reliability.}
Forecastability metrics such as spectral entropy, approximate entropy, and seasonality strength relate to signal difficulty \citep{tang2024timeseriesforecastingllms,wu2023timesnettemporal2dvariationmodeling,wang2025timeseriesforecastabilitymeasures,guntu2020wavelet}. While these metrics characterize data properties, they have not been systematically used to guide model selection at deployment time. 

\textbf{Our contribution} is not the $\Omega$ metric itself---spectral entropy is well-established---but rather the empirical discovery that zero-shot models exhibit a unique, systematic relationship with $\Omega$ that other model families do not. This differential response enables targeted model selection: for high-$\Omega$ data, the choice is clear; for low-$\Omega$ data, the advantages of these large models disappear. Table~\ref{tab:comparison} contrasts our approach with existing alternatives.

\begin{table}[h]
\centering
\caption{Comparison of model selection approaches. $\Omega$ uniquely provides model-family-specific guidance with minimal computation.}
\label{tab:comparison}
\small
\begin{tabular}{@{}lccc@{}}
\toprule
\textbf{Approach} & \textbf{Speed} & \textbf{Model Guidance} & \textbf{Interpretable} \\
\midrule
Spectral entropy (Wang 2025) & Fast & $\times$ & $\checkmark$ \\

% Approx. entropy \citep{pincus1991} & Fast & $\times$ & Medium \\
Approx. entropy \citep{pincus1991} & Fast & $\times$ & $\times$ \\

Validation subset & Medium & $\checkmark$ & $\times$  \\

Meta-learning (Talagal 2024) & Slow & $\checkmark$ & $\times$ \\

AutoML (Salehin 2024) & Very slow & $\checkmark$ & $\times$  \\

\midrule
\textbf{Spectral Predictability $\Omega$ (ours)} & \textbf{Fast} & \textbf{$\checkmark$} & \textbf{$\checkmark$} \\
\bottomrule
\end{tabular}

\end{table}

% \textbf{Model Selection in Practice.}
\textbf{Model Selection.}
Traditional model selection requires training and validating multiple candidates, which is resource-intensive. Meta-learning and AutoML approaches attempt to automate this process but still require significant computation \citep{li2025tsfm}. Our approach complements these methods by providing a fast preliminary filter based on data properties alone, enabling practitioners to focus expensive validation on a smaller subset of promising models.

% -------------------------------------------------------------------------
\section{Spectral Predictability~$\Omega$}
We quantify the inherent forecastability of a time series using spectral predictability $\Omega$, a metric grounded in information theory and signal processing.  
$\Omega$ captures how concentrated the energy is in the frequency domain: periodic series with strong seasonal patterns have concentrated spectra and high predictability, while noisy or irregular series yield diffuse spectra and low predictability~\citep{wang2025timeseriesforecastabilitymeasures, guntu2020wavelet}.

Let $\{x_t\}_{t=1}^T$ be a univariate series of length $T$. Apply a Hann taper and remove the DC component, then compute the FFT. Define the one-sided power spectral density:
\[
P_k = |\hat{x}_k|^2,\quad k=1,\dots,K,\quad K=\lfloor T/2 \rfloor,
\]
where $\hat{x}_k$ denotes the $k$-th frequency component (DC excluded). Normalize to obtain a probability distribution $p_k = P_k/\sum_{j=1}^K P_j$ and compute spectral entropy:
\[
H(x) = -\sum_{k=1}^K p_k \log p_k.
\]
Spectral predictability is defined by normalizing entropy by its maximum $H_{\max}=\log K$:
\[
\Omega(x)=1-\frac{H(x)}{H_{\max}}, \quad \Omega\in[0,1].
\]
High $\Omega$ indicates concentrated spectra (more predictable); low $\Omega$ indicates diffuse spectra (less predictable).

\textbf{Computational Efficiency.} Computing $\Omega$ requires only a single FFT pass, taking seconds on a standard laptop for typical forecasting datasets (thousands to millions of time points)---orders of magnitude faster than training even a single model. This makes $\Omega$ a practical preprocessing step for model selection.

\section*{Experimental Overview}

We assess spectral predictability~($\Omega$) through two stages:

\begin{itemize}
\item \textbf{Controlled Experiments:} Synthetic signals with tunable~$\Omega$ and three real datasets (CarbonCast, PEMS, Fitbit) test how forecasting error changes with $\Omega$. Models include TimeLLM (with both LLAMA3.2-1B and GPT2-130M backbones), randomly initialized backbone, and DLinear, evaluated by sMAPE and MSE.

\item \textbf{Large-Scale Analysis:} Using 51 models and 28 datasets from GIFT-Eval, we compute dataset-level $\Omega$ to compare statistical, deep-learning, pretrained, and zero-shot models across predictability levels.

\end{itemize}

\noindent
These experiments reveal how $\Omega$ captures forecasting difficulty and guides model selection.

% -------------------------------------------------------------------------
\section{Controlled Experiment Results: Establishing the Effect of $\Omega$}

To test whether spectral predictability genuinely affects forecasting difficulty—and can be systematically manipulated—we designed controlled experiments across four domains with varying characteristics:

\textbf{Synthetic Data.} We created synthetic Fourier signals explicitly engineered to span $\Omega$ values from 0.2 to 0.8. By controlling the spectral entropy directly through the frequency components, we generated time series with predetermined predictability levels.

\textbf{Real-World Domains.} We also tested three diverse real-world datasets: (i)~CarbonCast: hourly energy generation~\citep{maji2022carboncast}; (ii)~PEMS: hourly traffic flow~\citep{wang2024tssurvey}; and (iii)~Fitbit: minute-level heart rate~\citep{furberg_2016_53894}. These domains exhibit natural variation in $\Omega$ arising from different underlying processes, allowing us to verify that patterns observed in synthetic data generalize to realistic conditions.

\textbf{Models.} We evaluated four representative architectures: (i)~TimeLLM pretrained with frozen Llama3.2-1B weights; (ii)~the same architecture with random initialization; (iii)~GPT2-130M; and (iv)~DLinear~\citep{zeng2022transformerseffectivetimeseries,radford2019language}. All models used 512-step context and 96-step forecast horizon. Error was measured by the Symmetric Mean Absolute Percentage Error (sMAPE). sMAPE is a scale-normalized accuracy metric that lies in $[0,2]$ and is defined for a forecast $\hat{y}_t$ of target $y_t$ over $T$ timesteps as
\[
\text{sMAPE}
= \frac{1}{T} \sum_{t=1}^{T}
\frac{2 \lvert \hat{y}_t - y_t \rvert}{\lvert \hat{y}_t \rvert + \lvert y_t \rvert}.
\]
Lower values indicate better predictive accuracy, and because the denominator rescales by the magnitude of both the forecast and the ground truth at each timestep, sMAPE is comparable across datasets with different units and scales.

Further training details are in the Appendix.

\begin{figure}[t]
\centering
\begin{subfigure}{0.49\linewidth}
  \includegraphics[width=\linewidth]{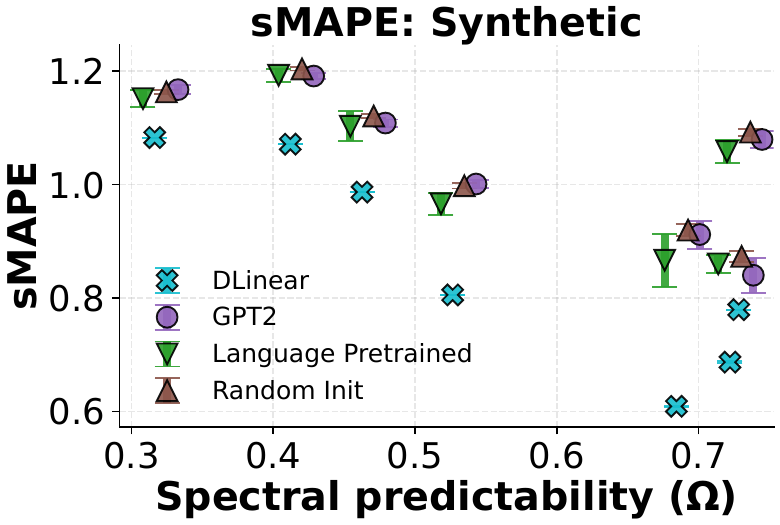}
  \caption{Synthetic}
\end{subfigure}
\begin{subfigure}{0.49\linewidth}
  \includegraphics[width=\linewidth]{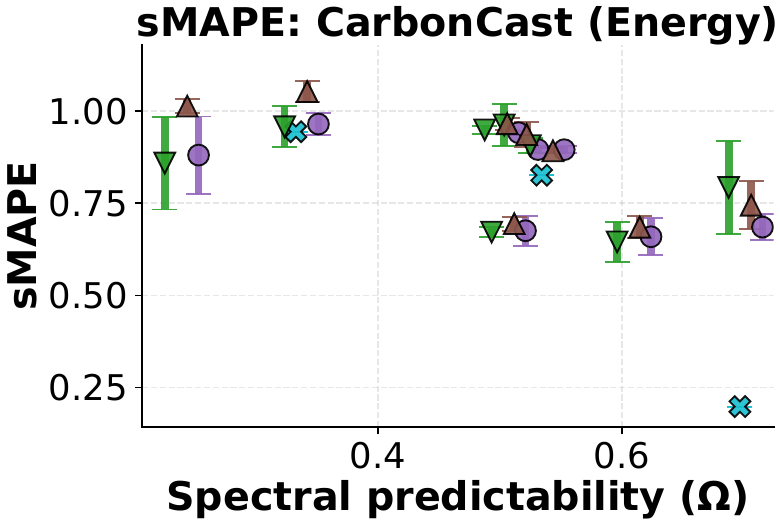}
  \caption{CarbonCast}
\end{subfigure}

\begin{subfigure}{0.49\linewidth}
  \includegraphics[width=\linewidth]{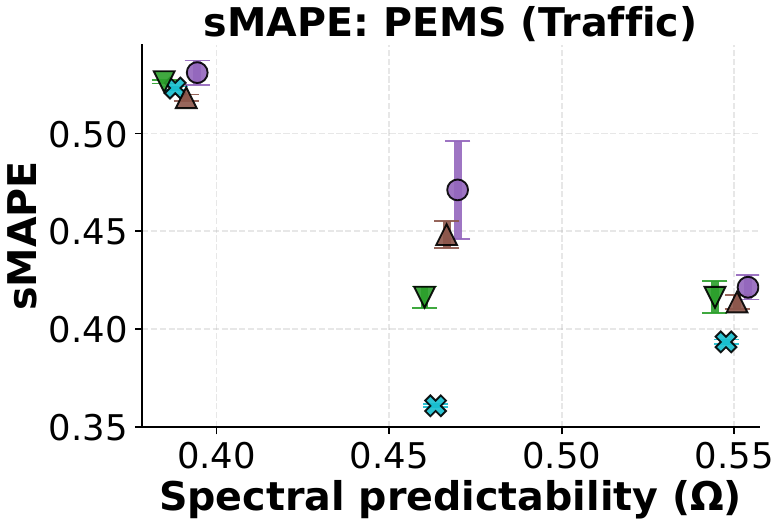}
  \caption{PEMS}
\end{subfigure}
\begin{subfigure}{0.49\linewidth}
  \includegraphics[width=\linewidth]{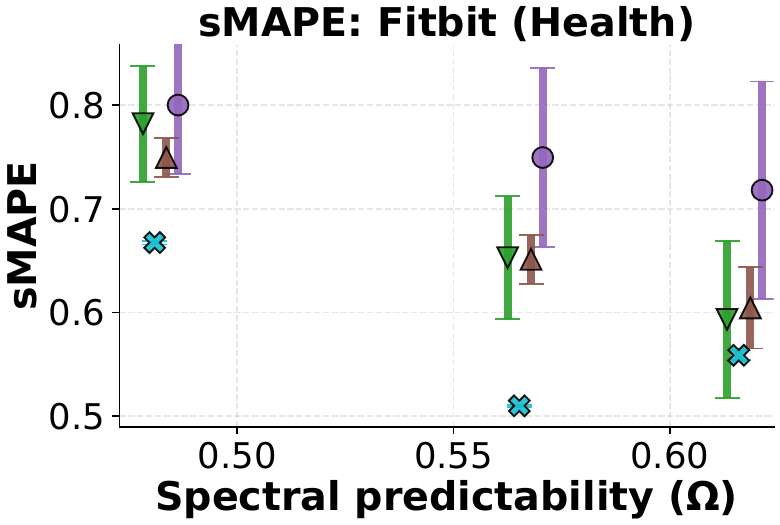}
  \caption{Fitbit}
\end{subfigure}

\caption{\textbf{Spectral predictability systematically affects forecasting difficulty.}
Across synthetic and real-world domains, sMAPE declines as $\Omega$ increases. Error bars show 95\% CIs across series. The clearest pattern emerges in synthetic data where $\Omega$ is directly controlled. Note that less data was available for PEMS and Fitbit, leading to sparser graphs. Also note that model classes have been slightly offset horizontally for visual clarity.}
\label{fig:base}
\end{figure}

\textbf{Consistency Across Metrics.} While our primary analysis uses sMAPE because it is normalized and allows for cross-dataset comparison, we verify the robustness of our observations using the popular MSE metric on controlled experiments. Table~\ref{tab:MSE_omega} shows the relationship between MSE and $\Omega$ exhibits consistent negative correlations across all domains (Pearson $r$ ranging from $-0.377$ to $-0.750$), confirming that the core pattern---error decreases as predictability increases---holds across error metrics. The consistency between sMAPE and MSE results suggests our findings are not artifacts of metric choice, though future work should examine probabilistic scores (CRPS, interval coverage) for additional validation.

\begin{table}[h]
\centering
\caption{Aggregate relationship between MSE and $\Omega$ across controlled experiments. Negative correlations indicate that forecasting error decreases as predictability increases, supporting $\Omega$ as a proxy for difficulty.}
\label{tab:MSE_omega}
\begin{tabular}{lrr}
\toprule
Dataset & Pearson $r$ & Spearman $\rho$ \\
\midrule
Synthetic         & $-0.720$ & $-0.678$ \\
CarbonCast           & $-0.676$ & $-0.740$ \\
PEMS              & $-0.750$ & $-0.708$ \\
Fitbit & $-0.377$ & $-0.367$ \\
\bottomrule
\end{tabular}
\end{table}

\textbf{Key Findings.}
Fig.~\ref{fig:base} shows noticeable patterns across all domains: forecasting error systematically decreases as $\Omega$ increases. This effect is most pronounced in Synthetic, where we engineered $\Omega$ directly, providing strong evidence that spectral predictability correlates with difficulty. The pattern replicates in CarbonCast (energy) and, to a lesser extent, in PEMS (traffic) and Fitbit (wearables).

In Synthetic and CarbonCast, where spectral structure dominates signal characteristics, the $\Omega$-error relationship is nearly monotonic. Models tend to show improved performance at high $\Omega$, with error reductions of 20--40\% when moving from $\Omega=0.3$ to $\Omega=0.7$.

The effect is weaker in PEMS and Fitbit, likely because other factors—missingness patterns (Fitbit users removing devices), noise characteristics, and domain-specific irregularities—contribute substantially to difficulty beyond spectral properties alone. This suggests $\Omega$ is a useful but not exhaustive indicator; practitioners should consider it alongside domain knowledge.

These controlled experiments suggest a key result: spectral predictability systematically stratifies forecasting difficulty. However, these experiments lack scale and leave open a critical question for practitioners: \textbf{do different model families—statistical, deep learning, pretrained or TSFM—respond differently to $\Omega$?} Understanding this would enable targeted model selection based on dataset properties. We investigate this next in a more comprehensive setting.

\section{Large-Scale Analysis Results: Model-Family-Specific Responses to $\Omega$}
\label{sec:large-scale}
To examine whether different model families exhibit distinct relationships with spectral predictability, we analyzed 51 models from the \textbf{GIFT-Eval Time Series Forecasting Leaderboard}, using their reported sMAPE performance across 28 datasets spanning energy, healthcare, finance, and natural domains~\citep{aksu2024giftevalbenchmarkgeneraltime}. Each model was categorized as \emph{statistical}, \emph{deep-learning}, \emph{pretrained}, or \emph{zero-shot} following GIFT-Eval's taxonomy. Further model type categories include \emph{fine-tuned} and \emph{agentic}, though they are not the focus of this study due to the small number of representatives at the time of writing. All models used and their respective categories are reported in the Appendix. In this context, both pretrained and zero-shot models are large TSFM models applied directly without fine-tuning. However, certain models (eg. TimesFM) were originally trained with some amount of data that is in the GIFT-Eval evaluation dataset. To prevent leakage, these models were then \emph{pretrained} on a leak-free dataset designed by~\citep{aksu2024giftevalbenchmarkgeneraltime} and are considered ``pretrained". On the other hand, models labeled as ``zero-shot" (such as TimesFM-2.5) are TSFMs with no data leakage in their published weights and thus not ``re-pretrained".

 Train--test splits were not public, so we computed $\Omega$ over each full dataset to characterize its overall spectral properties. This aggregate $\Omega$ serves as a dataset-level descriptor that does not inform individual predictions. Our goal was to identify systematic patterns in how different model families respond to varying levels of predictability.

\begin{figure}[t!]
\centering
\includegraphics[width=\linewidth,height=0.30\textheight,keepaspectratio]{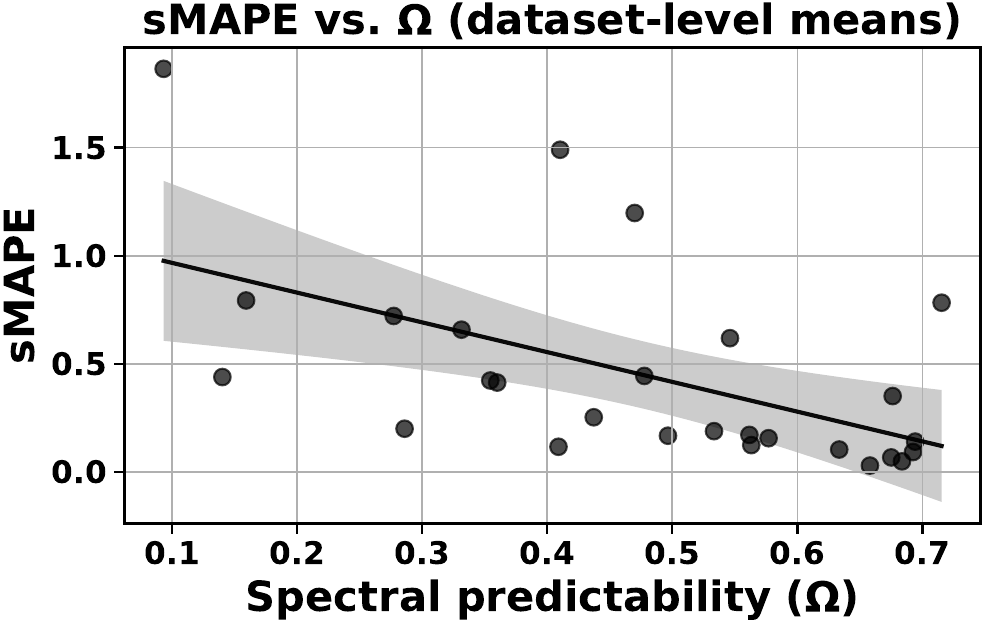}
\caption{\textbf{Predictability-error relationship at scale.} Across 28 datasets and 51 models, average error (sMAPE) declines with increasing spectral predictability~$\Omega$. Each point represents an average of all models on one dataset. We fit an ordinary least squares line of best fit with 95$\%$ confidence interval for visualization.}
\label{fig:gifteval_overall}
\end{figure}

\begin{figure}[t!]
\centering
\includegraphics[width=\linewidth,height=0.30\textheight,keepaspectratio]{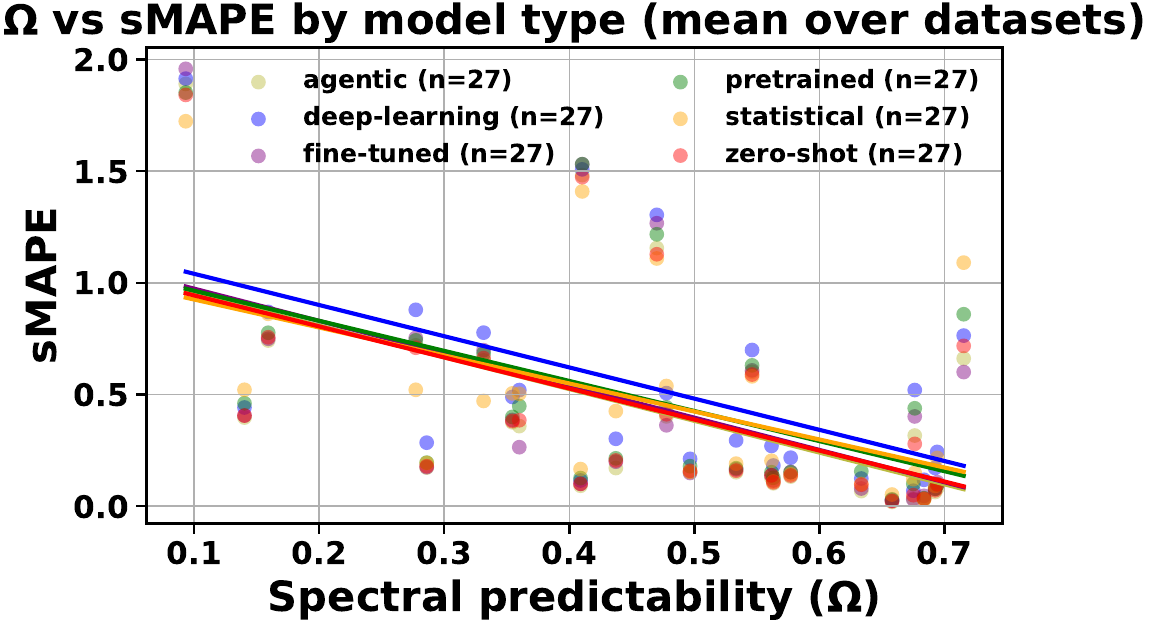}
\caption{\textbf{Predictability-error relationship with model types split out.} The model type classes were taken from GIFT-Eval's classification~\citep{aksu2024giftevalbenchmarkgeneraltime}.}
\label{fig:gifteval_bymodeltype}
\end{figure}

\begin{figure}[t!]
\centering
\includegraphics[width=\linewidth,height=0.30\textheight,keepaspectratio]{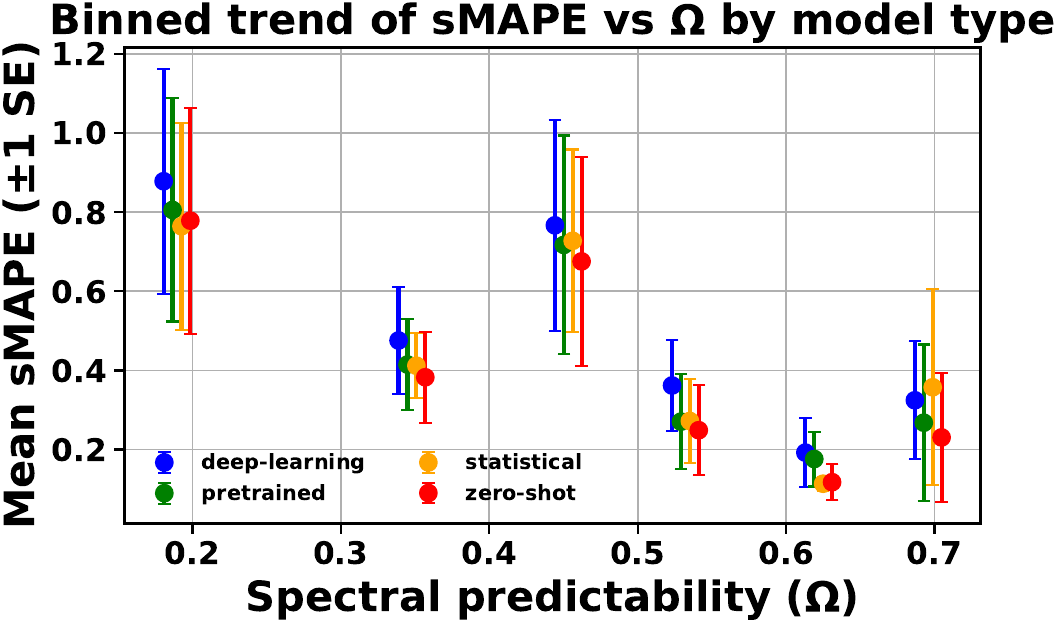}
\caption{\textbf{Model Types Binned for Clarity.} Only model types with more than 4 representative models were chosen to be represented here for robustness and visual clarity.}
\label{fig:gifteval_binnedbymodeltype}
\end{figure}

\textbf{Overall Pattern.}
Across the 28 datasets, we found a statistically significant monotonic relationship between predictability and error (Spearman $\rho=-0.65$, $p=1.9\times10^{-21}$), confirming that the pattern observed in controlled experiments generalizes at scale (Fig.~\ref{fig:gifteval_overall}). Results for sMAPE versus $\Omega$ by model type are presented in Fig.~\ref{fig:gifteval_bymodeltype}, and as binned averages in Fig.~\ref{fig:gifteval_binnedbymodeltype}, suggesting that this trend is consistent for different model types.
To produce the bins in Fig. \ref{fig:gifteval_binnedbymodeltype}, datasets are grouped into 6 quantile bins of $\Omega$, which ensures that each bin contains a similar number of datasets, preventing high-density $\Omega$ regions from dominating the analysis. Each plotted point represents the averaged sMAPE for one of the 4 given model classes in a given $\Omega$ regime, with vertical error bars showing the uncertainty across datasets. Model classes are slightly offset horizontally within each bin to avoid overlap and improve visual separation.

\subsection{Relationship with Chaos (Largest Lyapunov Exponent)}

To investigate whether spectral predictability correlates with chaotic dynamics, we computed the Largest Lyapunov Exponent (LLE) for each dataset. The LLE measures a system's sensitivity to initial conditions: it quantifies the average exponential rate at which two nearby trajectories in the reconstructed state space diverge. Formally,
\[
\lambda_{\max}
= \lim_{\Delta t \to \infty}
\frac{1}{\Delta t} \;
\Big\langle
\ln \frac{\lVert \delta \mathbf{x}(t + \Delta t) \rVert}{\lVert \delta \mathbf{x}(t) \rVert}
\Big\rangle,
\]
where $\delta \mathbf{x}(t)$ is an infinitesimal perturbation between two initially close states of the same sequence. Higher LLE indicates more chaotic, less locally predictable dynamics.

\begin{figure}[t!]
\centering
\includegraphics[width=0.7\linewidth,keepaspectratio]{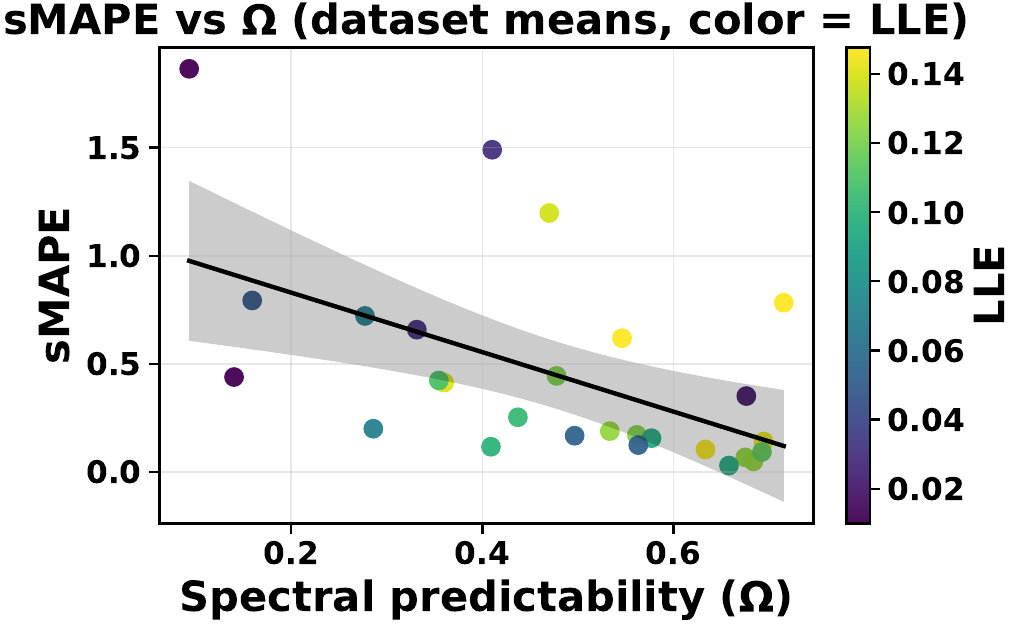}
\caption{\textbf{Relationship with LLE.} A heatmap showing LLE and Omega. A higher LLE corresponds to larger amounts of chaos in a sequence. This plot implies that some of the datasets which exhibit high $\Omega$ also have high chaos, which could explain why the sMAPE unexpectedly worsens.}
\label{fig:omega_lle}
\end{figure}

Fig.~\ref{fig:omega_lle} shows a counterintuitive pattern: datasets with higher $\Omega$ (more predictable spectra) sometimes exhibit higher LLE values (suggesting more chaos). This apparent paradox arises because spectral predictability and dynamical chaos measure different aspects of time series structure. $\Omega$ captures frequency-domain regularity (periodic or quasi-periodic patterns), while LLE measures sensitivity to initial conditions in phase space. A series can have highly structured spectral content (high $\Omega$) while still being chaotic in the deterministic sense. Importantly, this complexity indicates that while higher $\Omega$ associates with lower forecasting error, other qualities of the dataset can also have an impact and deserve further investigation.

\textbf{Model-Family-Specific Patterns.}
To examine differential responses, we compared relative accuracies between model types for each dataset. For each model pair $A\!\to\!B$ evaluated on dataset~$i$:
\[
\Delta_{A \rightarrow B}^{\text{sMAPE}}(i) =
100 \times \frac{\mathrm{sMAPE}(A,i) - \mathrm{sMAPE}(B,i)}{\mathrm{sMAPE}(A,i)}.
\]
Negative $\Delta$ indicates Model~$A$ achieves lower error (better performance) than Model~$B$.

The solid red curve is a LOWESS (locally weighted scatterplot smoothing) fit of $\Delta$($A$ $\rightarrow$ $B$) as a function of $\Omega$, using a smoothing fraction of 0.4. The shaded red region is an empirical 95$\%$ confidence band for that trend, constructed via a nonparametric bootstrap: we resample datasets with replacement 300 times, recompute the LOWESS fit for each bootstrap resample, and evaluate each fitted curve on a common ($\Omega$) grid. For each ($\Omega$) value on that grid, we take the 2.5th and 97.5th percentiles across the bootstrap fits; these percentiles define the lower and upper edges of the shaded band. Thus, the solid red line is a visualizer that shows the smoothed observed relationship between ($\Omega$) and relative error gain, and the translucent band shows the bootstrap variability of that relationship across datasets.

\begin{figure}[t!]
\centering
\begin{subfigure}{0.49\linewidth}
\includegraphics[width=\linewidth,height=0.18\textheight,keepaspectratio]{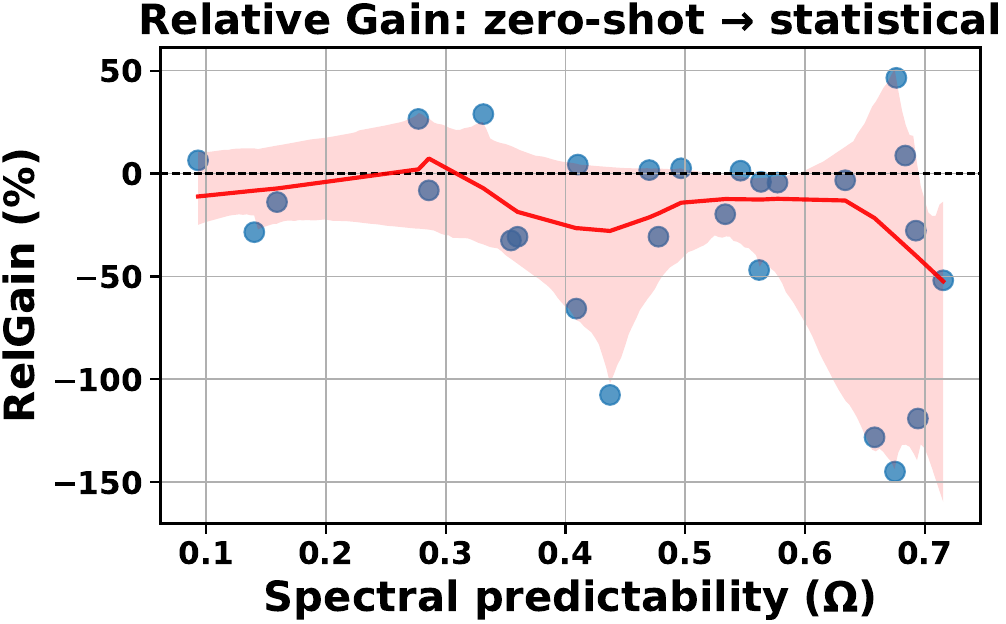}\\[0.7em]
\end{subfigure}
\begin{subfigure}{0.49\linewidth}
\includegraphics[width=\linewidth,height=0.18\textheight,keepaspectratio]{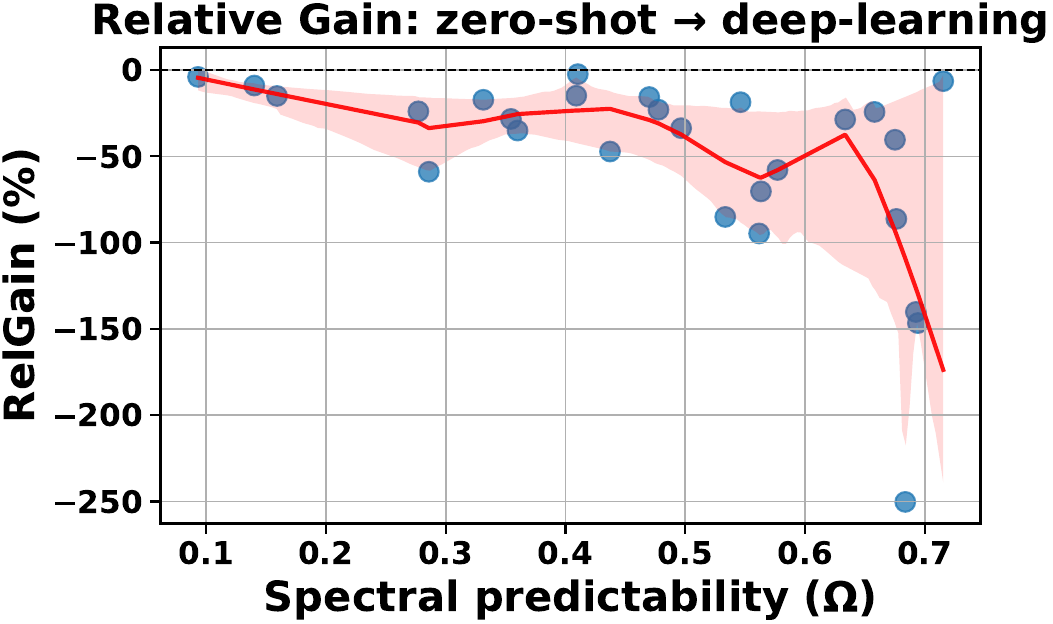}\\[0.7em]
\end{subfigure}
\begin{subfigure}{0.49\linewidth}
\includegraphics[width=\linewidth,height=0.18\textheight,keepaspectratio]{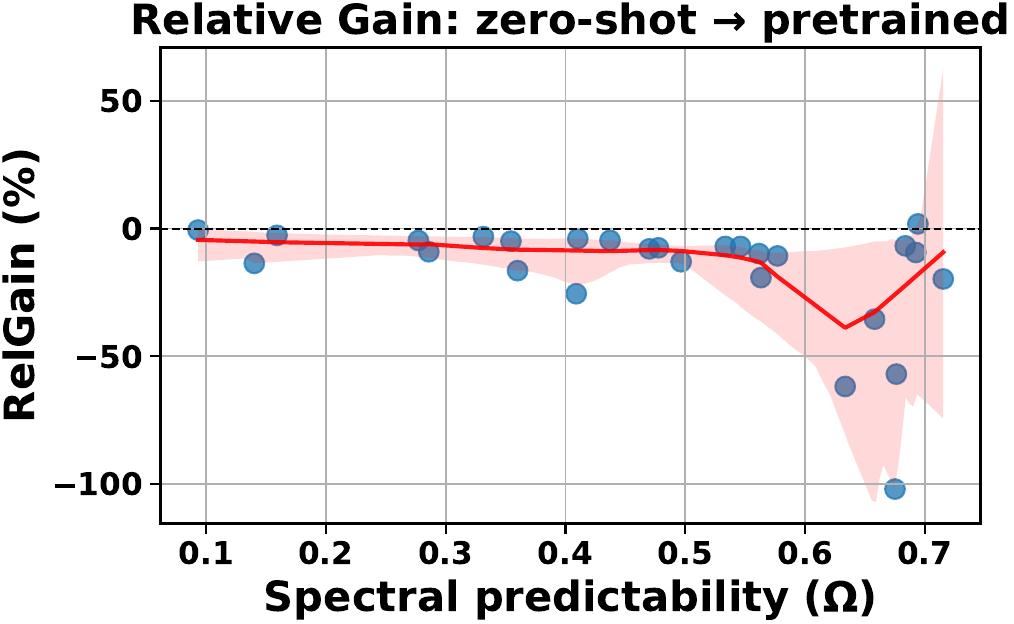}\\[0.2em]
\end{subfigure}
\caption{\textbf{Zero-shot models uniquely exploit high-$\Omega$ regimes.}
Negative values indicate zero-shot models achieve lower error. The average performance advantage improves with $\Omega$.}
\label{fig:zeroshotRel}
\end{figure}

Fig.~\ref{fig:zeroshotRel} reveals a striking pattern: \textbf{zero-shot models show systematically increasing advantages as $\Omega$ increases}, with performance gains reaching 20--60\% over statistical and deep learning baselines in high-$\Omega$ regimes ($\Omega > 0.5$). This relationship is moderately consistent across all three comparisons (vs.~statistical, vs.~deep learning, vs.~pretrained), with Spearman correlations ranging from -0.234 for statistical to -0.556 for deep learning.

\begin{figure}[t!]
\centering
\begin{subfigure}{0.49\linewidth}
\includegraphics[width=\linewidth,height=0.18\textheight,keepaspectratio]{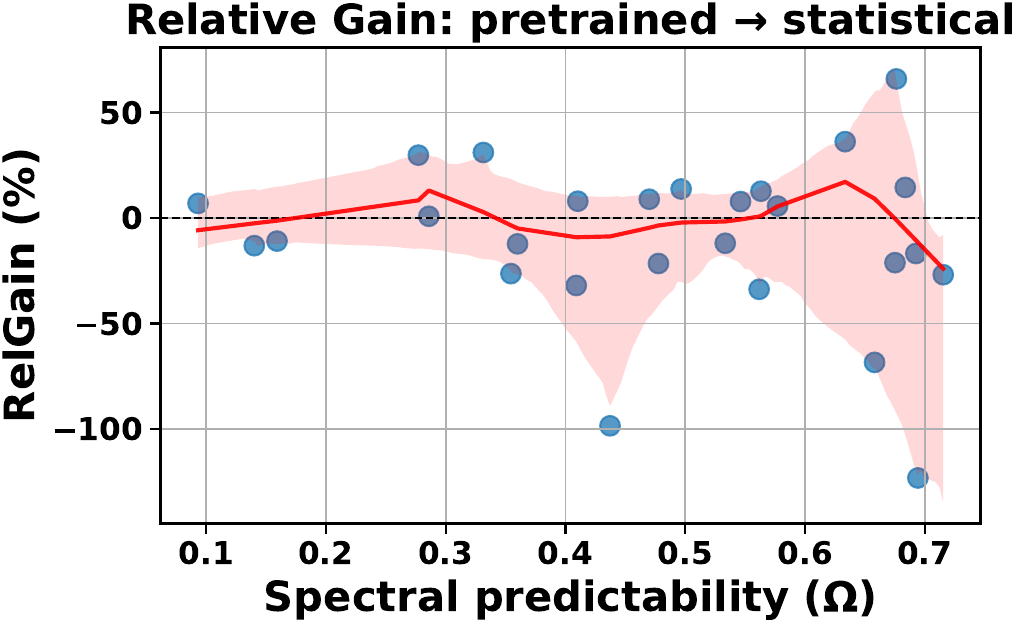}\\[0.7em]
\end{subfigure}
\begin{subfigure}{0.49\linewidth}
\includegraphics[width=\linewidth,height=0.18\textheight,keepaspectratio]{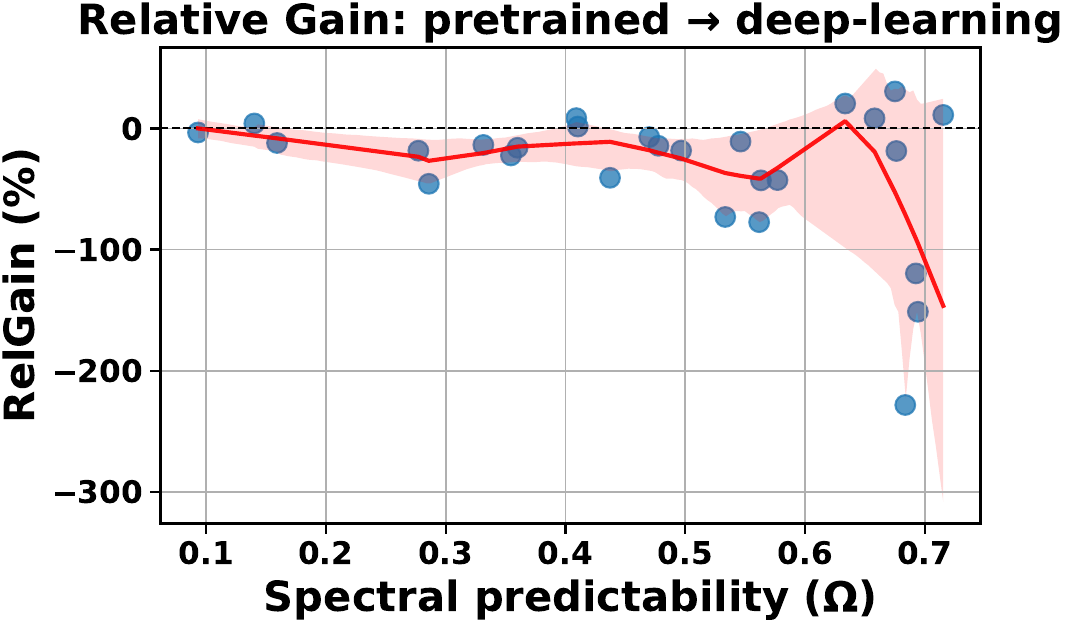}\\[0.2em]
\end{subfigure}

\caption{\textbf{Pretrained models show no systematic $\Omega$-dependence.}
Unlike zero-shot models, pretrained models do not exhibit predictable advantages based on spectral predictability.}
\label{fig:pretrainRel}
\end{figure}

Critically, this advantage is \emph{specific to zero-shot models}. Fig.~\ref{fig:pretrainRel} shows that pretrained models exhibit no consistent $\Omega$-dependent pattern when compared to statistical or deep learning baselines. Their relative performance displays minimal trend across the $\Omega$ spectrum, suggesting that the process of ``re-pretraining" TSFMs somehow degraded their ability to exploit the highest-$\Omega$ ranges. This suggests that the pretraining corpus—not just model architecture or 
size—fundamentally shapes how models respond to spectral structure, a promising future area of investigation.

\textbf{The Low-$\Omega$ Challenge.}
Critically, as $\Omega$ decreases below 0.2, performance differences between models narrow substantially. At $\Omega < 0.2$, there is minimal differentiation between between model types; all struggle similarly. This reveals an important research gap: current models, whether lightweight or sophisticated, have made limited progress on genuinely difficult (low-$\Omega$) problems. The field has optimized performance on predictable data without developing architectures that push advantages where forecasting is hardest.

\subsection{Practical Implications for Model Selection}

These results provide clear, actionable guidance for practitioners:

\textbf{High-$\Omega$ Datasets ($\Omega > 0.5$):} Zero-shot models are the optimal choice. A practitioner who computes $\Omega$ and finds their dataset falls in this regime can narrow their search space to zero-shot candidates (e.g., Moirai or Chronos) rather than validating a multitude of diverse models. This reduces the computational costs of validation substantially while maintaining or improving expected performance.

\textbf{Low-$\Omega$ Datasets ($\Omega < 0.4$):} Zero-shot models provide no clear advantage. Simpler statistical or deep learning models (like ARIMA or DLinear, respectively) offer comparable accuracy with substantially lower computational cost (both training and inference). For resource-constrained deployments or when rapid iteration is needed, this tradeoff strongly favors lightweight alternatives.

\textbf{Mid-$\Omega$ Datasets ($0.4 \leq \Omega \leq 0.5$):} The choice is less clear-cut. Practitioners should consider additional factors: computational budget, inference latency requirements, and whether the dataset exhibits characteristics (missingness, regime shifts) that $\Omega$ may not fully capture.

\subsection{When to Trust $\Omega$: Reliability Conditions}

While $\Omega$ provides useful guidance across diverse datasets, its reliability depends on specific data characteristics. We identify three conditions that limit $\Omega$'s predictive power:

\textbf{(1) Non-stationary processes.} $\Omega$ assumes spectral properties remain stable over time. For series with regime shifts or structural breaks, or those with high LLE, a single aggregate $\Omega$ may not reflect local forecasting difficulty.

\textbf{(2) Insufficient data length.} FFT-based spectral estimates become unstable for short series (we recommend $T > 1000$ timesteps).

\textbf{(3) Exogenous-driven dynamics.} $\Omega$ characterizes intrinsic temporal structure. For processes dominated by external shocks (e.g., traffic accidents, policy interventions), spectral properties may not capture true predictability. The weaker patterns in PEMS and Fitbit (Fig.~\ref{fig:base}) likely reflect such factors.

\textbf{Failure case example.} In Fig.~\ref{fig:zeroshotRel}, an outlier dataset point exhibits $\Omega=0.66$ but statistical models outperform zero-shot by nearly 50\%, counter to expectations. Further analysis is needed to understand dynamics not fully captured by $\Omega$.

\textbf{Decision heuristic.} Before applying $\Omega$-based selection, practitioners should verify: (i)~series length is sufficient for a rigorous computation of $\Omega$, (ii)~visual inspection or statistical tests suggest stationarity, and (iii)~domain knowledge indicates intrinsic temporal patterns dominate exogenous shocks. When these conditions hold, $\Omega$ provides reliable guidance; otherwise, complement with domain-specific heuristics.

%-------------------------------------------------------------------------
\section{Discussion}

\subsection{Why Not Just Use Validation Error?}

A natural question arises: why not simply train all candidate models and pick the one with lowest validation error? We argue spectral predictability offers three key advantages:

\textbf{(1) Computational Efficiency.}
Computing $\Omega$ takes seconds; training and validating multiple models can take hours to days. For a practitioner with many candidate models, our approach reduces the search space \emph{before any training begins}, saving substantial compute and engineering time. This represents meaningful cost savings, especially when iterating across multiple datasets or deployment scenarios.

\textbf{(2) Interpretability and Generalization.}
Validation error reports \emph{which} model performed best on a specific dataset, but not \emph{why}. $\Omega$ provides interpretable signal properties that generalize across datasets. If a practitioner encounters a new high-$\Omega$ dataset, they can leverage prior knowledge that zero-shot models excel in this regime without re-running exhaustive validation. This builds intuition and enables faster decision-making.

\textbf{(3) Highlighting Research Gaps.}
Validation error alone doesn't reveal systematic patterns in model behavior. By stratifying performance along $\Omega$, we expose that models primarily differentiate on easy problems (high-$\Omega$) while converging on hard ones (low-$\Omega$). This insight motivates targeted research into improving performance where it matters most.

In practice, we envision $\Omega$ as a \emph{first-pass filter} that complements, rather than replaces, validation. The decision process: compute $\Omega$, narrow the model space based on regime, then validate the reduced candidate set. This hybrid approach balances efficiency with empirical rigor.

\subsection{Limitations}

\textbf{Other Correlational Effects.} Though our exploration makes a case for $\Omega$ as a predictor of model performance, our preliminary results suggest possible effects on accuracy from both the pretraining corpus of the TSFM selected and the LLE of the data. More work should be done to integrate these indicators into a more cohesive index.

\textbf{Model coverage.} GIFT-Eval provides broad coverage but may not represent all architecture types. Notably absent are neural ODEs and certain probabilistic models that might exhibit different $\Omega$ relationships.

\textbf{Aggregation effects.} Computing $\Omega$ on full datasets when train-test splits are unavailable creates potential for subtle aggregation bias. Our controlled experiments (where $\Omega$ is computed per-series on test data only) replicate key patterns, suggesting findings are robust, but future work should examine sensitivity to this choice.

% -------------------------------------------------------------------------
\section{Conclusion and Future Work}

Our key findings provide actionable guidance: for high-$\Omega$ datasets ($\Omega > 0.5$), zero-shot models generally outperform alternatives; for low-$\Omega$ datasets ($\Omega < 0.4$), zero-shot advantages disappear and lightweight models offer comparable accuracy at lower cost. We identify when $\Omega$ is reliable (stationary processes, sufficient length) and when practitioners should complement it with domain knowledge (regime shifts, exogenous shocks).

Beyond practical model selection, our analysis reveals that current models differentiate primarily on easy (high-$\Omega$) problems, with performance gaps narrowing substantially as $\Omega$ decreases. This highlights a critical research need: developing architectures that excel specifically on genuinely difficult forecasting challenges. We propose specific directions including attention mechanisms for sparse pattern detection, mixture-of-experts for adaptive capacity allocation, and explicit noise modeling.

Future work can extend this framework along several directions.
First, $\Omega$ could play a central role in \emph{agentic time-series AI}: autonomous systems that monitor their data streams, estimate $\Omega$ in real time, and self-configure by choosing lightweight or foundation-scale models as conditions change. 
Second, integrating $\Omega$ as a standardized metadata field in future benchmarks would promote reproducibility and allow systematic comparisons of model reliability across predictability regimes.
Third, our findings highlight the need to advance modeling in the \emph{low-$\Omega$ regime}, where current architectures converge in performance.
Finally, generalizing ($\Omega$) to \emph{multivariate and multimodal} time series would enable a notion of joint predictability that captures more complex dependencies.

Together, these directions outline a vision of predictability-aware AI systems that use spectral structure not only to guide model selection but also to drive self-adaptive, resource-efficient, and interpretable time-series intelligence.

\section{Acknowledgements}
The research reported in this paper was sponsored in part by the National Science Foundation (awards $\#$ CNS-2211301 and CNS-232595), the National Institutes of Health (award $\#$ 1P41EB028242), and the DEVCOM Army Research Laboratory (award $\#$ W911NF1720196). The views and conclusions contained in this document are those of the authors and should not be interpreted as representing the official policies, either expressed or implied, of the funding agencies.
Mani Srivastava was also partially supported by the Mukund Padmanabhan Term Chair at UCLA.

% -------------------------------------------------------------------------
\bibliographystyle{plainnat}
\bibliography{refs}

\appendix
\section{Appendix}

\subsection{Model Categorization by Type}
\label{app:model_types}

For completeness, we list below which models fall under each model type used in our analyses. Note that similar models in the same family may have been applied differently (e.g., used with pretraining versus in zero-shot). The models labeled as ``agentic" have an emphasis on dynamic ensemble models.

\textbf{GIFT-Eval Zero-shot models:}
TimesFM-2.5, FlowState-9.1M, Kairos\_10m, Kairos\_23m, Kairos\_50m, sundial\_base\_128m, YingLong\_110m, YingLong\_300m, YingLong\_50m, YingLong\_6m, granite-flowstate-r1, TiRex, DeOSAlphaTimeGPTPredictor-2025, Toto\_Open\_Base\_1.0, TabPFN-TS, Moirai\_base, Moirai\_large, Moirai\_small, and VisionTS.
\textbf{Pretrained models:}
Moirai2, Chronos\_bolt\_base, Chronos\_bolt\_small, Chronos\_large, Chronos\_base, Chronos\_small, TimesFM, timesfm\_2\_0\_500m, TTM-R1-Pretrained, TTM-R2-Pretrained, and Lag-Llama.
\textbf{Deep learning models:}
xLSTM-Mixer, PatchTST, TFT, N-BEATS, DLinear, TIDE, DeepAR, Crossformer, and iTransformer.
\textbf{Statistical baselines:}
Auto\_Arima, Seasonal\_Naive, Auto\_Theta, Auto\_ETS, and Naive.
\textbf{Fine-tuned models:}
TSOrchestra-test, TTM-R2-Finetuned, and TEMPO\_ENSEMBLE.
\textbf{Agentic models:}
Kairos-1.0, Nexus-1.0, TSOrchestra, and TimeCopilot.

\subsection{Dataset Protocol}
\label{app:data}
\textbf{Cross-series generalization.}
Models are trained on heterogeneous series within each domain, spanning a wide range of spectral entropies, and evaluated on held-out series covering similar ranges. This enables assessment of how well models generalize to unseen series with varying $\Omega$ values.

For real-world datasets, we select training series to represent the full range of spectral entropies available in each domain.  
CarbonCast and PEMS are binned into low, medium, and high tertiles, with 6 series randomly chosen from each bin (18 total).  
For Fitbit, which has fewer users, we select 5 series at each quintile.  
Test data come from held-out series at the 5th, 50th, and 95th percentiles of spectral entropy. Note that this explains why the entropy range within each figure's subplots may vary. 

Synthetic training data sweep 8 spectral entropy levels from 0.25 to 0.85 in increments of 0.10; test data cover 0.2 to 0.8 with the same intervals.  

PEMS and CarbonCast originate as multivariate datasets.  
PEMS is aggregated to hourly resolution (following \citet{wang2024tssurvey}), and all covariates are split into independent univariate series to enable spectral entropy computation and avoid covariate learning effects.  
These series are treated as same-domain data with distribution shifts (eg. across regions in PEMS or energy sources in CarbonCast).  
For Fitbit, user series with missing (zero) values in the test window are excluded.  

\textbf{GIFT-Eval Conversion}
For the original datasets present in GIFT-Eval's analysis, we calculated metrics such as $\Omega$ and LLE per each covariate, and averaged them together for a multivariate dataset. Due to runtime constraints and some very large datasets, we truncated the incoming data at 4096 steps per data stream. To calculate LLE, we reconstructed the state space with an embedding dimension $m=4$ and delay $\tau=10$ steps. We then estimated the LLE using standard Rosenstein-style local divergence tracking. These hyperparameters were chosen to ensure sufficient samples for a robust estimate without taking prohibitively long, and further investigation can be done here. 

\subsection{Training Protocol}
\label{app:train}
Single H100 GPU, three seeds per model. LR~$=0.01$, batch~$=16$, up to 10 epochs with early stopping (patience 3). Inputs normalized per context window; stride 1; boundary regime prevents window crossing across different series.

\subsection{Spectral Predictability Details}
\label{app:spectral}
We apply a Hann taper window for a balanced compromise between main-lobe width and side-lobe suppression to reduce spectral leakage. We choose this because we analyze diverse time series whose periodicities are not known in advance. For controlled experiments, we compute $\Omega$ from each test series (using only the input sequence, not forecast targets) to characterize the spectral properties each model must forecast. This avoids leakage while providing a per-series difficulty measure.

\subsection{Statistical Results of Controlled Experiments}
\label{app:stats}
\begin{table}[h]
\small
%\resizebox{\linewidth}{!}{
\centering
\caption{Correlation statistics between $\Omega$ (spectral predictability) and sMAPE. 95\% CIs computed via Fisher $z$-transform. Narrow intervals reflect low variability across seeds.}
\label{tab:corr_omega_smape_transposed}
\begin{tabular}{lcccc}
\toprule
\textbf{Statistic} & \textbf{CarbonCast} & \textbf{Fitbit} & \textbf{PEMS} & \textbf{Synthetic} \\
\midrule
$n$                  & 33  &  18  &  18  & 42 \\
Pearson $r$          & $-0.68$ & $-0.38$ & $-0.75$ & $-0.72$ \\
95\% CI (low)        & $-0.83$ & $-0.72$ & $-0.90$ & $-0.84$ \\
95\% CI (high)       & $-0.43$ & $-0.11$ & $-0.43$ & $-0.53$ \\
Spearman $\rho$      & $-0.74$ & $-0.38$ & $-0.71$ & $-0.68$ \\
\bottomrule
\end{tabular} 

\end{table}

In addition to our controlled experiment conclusions identified earlier, we report the corresponding tabular results in MSE, though similar trends hold for sMAPE.  
Aggregate statistics are computed as the mean of the domain-specific values for clarity and comparability.  

\textbf{(i)~Forecasting Error Decreases as Predictability Increases.}  
Table~\ref{Tab Base Slope} reports the aggregate relationship between MSE and spectral predictability ($\Omega$) across models.  
Across all settings, the median slopes for MSE are also negative, suggesting that error systematically declines as predictability rises.  
This trend holds for lightweight models such as DLinear as well as for larger pretrained backbones, reinforcing $\Omega$ as a proxy for forecasting difficulty.  
Notably, DLinear shows the smallest magnitude of slope, reflecting its strong performance in predictable regimes, while language-pretrained models and random initialization exhibit slightly steeper slopes, indicating greater sensitivity to $\Omega$.  

\begin{table}[t]
\centering
\caption{Aggregate relationship between $\Omega$ and MSE. Negative slopes indicate that forecasting error decreases as predictability increases, supporting $\Omega$ as a proxy for difficulty.}
\label{Tab Base Slope}
\begin{tabular}{lrr}
\toprule
Model & Median slope \\
\midrule
Aggregate         & -1.08  \\
DLinear           & -0.97 \\
GPT2              & -1.08 \\
Language Pretrained & -1.17 \\
Random Init       & -1.09 \\
\bottomrule
\end{tabular}
\end{table}

\textbf{(ii)~DLinear Dominates in Predictable Regimes.} Table~\ref{LangDlin} presents the Theil–Sen slope of the relative Error Increase $\Delta$ (\%) versus $\Omega$, comparing Language Pretrained models against DLinear.  
Positive slopes indicate that the performance gap in favor of DLinear widens as predictability increases, suggesting  a strong inductive bias for structured series.  
The table also reports Spearman $\rho$ and Pearson $r$ correlations to assess robustness.  
Results show consistent positive slopes in CarbonCast, PEMS, and Synthetic, with strong correlations ($r > 0.85$), confirming that DLinear outperforms pretrained models in high-$\Omega$ regimes. 
The aggregate trend is likewise positive, while Fitbit deviates with a negative slope and weak correlations, highlighting the unique irregularities of wearable data.  
Overall, these results reinforce that DLinear is most effective when time series exhibit strong seasonality or repetition, while pretrained models gain relevance as predictability falls.

\begin{table}[t]
\centering
\caption{Theil--Sen slope of Error Increase $\Delta$ (\%) vs $\Omega$ for Language Pretrained relative to DLinear. Positive slopes indicate that performance gap in favor of DLinear widens as predictability increases.}  
\label{LangDlin}
\begin{tabular}{lrrr}
\toprule
Domain & Slope & Spearman $\rho$ & Pearson $r$ \\
\midrule
Aggregate & 87.97 & 0.57 & 0.58 \\
CarbonCast & 225.41 & 1.00 & 0.95 \\
Fitbit     & -47.23 & -0.50 & -0.47 \\
PEMS       & 72.73  & 1.00 & 0.96 \\
Synthetic  & 100.97 & 0.79 & 0.87 \\
\bottomrule
\end{tabular}
\end{table}

\textbf{(iii)~Model Size Has Limited Effect Across Predictability Levels.}  
Table~\ref{LangGPT} is structured similarly to Table~\ref{LangDlin}, but compares Language Pretrained models to GPT-2 (a smaller backbone).  
It evaluates how model size influences relative performance when conditioned on spectral predictability $\Omega$.  
The aggregate slope is near zero ($-2.08$), indicating little systematic advantage for either model overall.  
CarbonCast shows a strongly positive slope ($69.99$) with high correlations ($r=0.94$), suggesting that larger pretrained models perform better on high-$\Omega$ energy data.  
In contrast, Fitbit again shows a strongly negative slope ($-88.34$), reflecting its irregular missingness and confirming it as an exception.  
PEMS and Synthetic display small, weakly positive slopes with low correlations, reinforcing that scaling effects are inconsistent across domains.  
Overall, these results suggest that while model size may provide gains in certain settings, $\Omega$ remains the dominant axis for explaining relative performance, and scaling alone does not guarantee improvements across all domains.  

\begin{table}[t]
\centering
\caption{Theil--Sen slope of Error Increase $\Delta$ (\%) versus $\Omega$ for Language Pretrained relative to GPT-2, showing how model size affects performance across predictability levels.}  
% \caption{Theil--Sen slope of Error Increase $\Delta$ (\%) vs. $\Omega$ for Language Pretrained relative to GPT-2.}
\label{LangGPT}
\begin{tabular}{lrrr}
\toprule
Domain & Slope & Spearman $\rho$ & Pearson $r$ \\
\midrule
Aggregate & -2.08 & 0.32 & 0.10 \\
CarbonCast & 69.99 & 1.00 & 0.94 \\
Fitbit     & -88.34 & -0.50 & -0.87 \\
PEMS       & 7.27   & 0.50 & 0.19 \\
Synthetic  & 2.76   & 0.29 & 0.15 \\
\bottomrule
\end{tabular}
\end{table}

\textbf{(iv)~Pretraining Provides Limited Gains Beyond the Embedding Head.}  
Table~\ref{LangRand} follows the structure of Table~\ref{LangDlin}, but compares Language Pretrained models to Random Init, isolating the effect of pretraining.  
It evaluates how pretraining influences relative performance across levels of spectral predictability $\Omega$.  
The aggregate slope is small ($15.8$) with negligible correlations, suggesting limited systematic benefit of pretraining overall.  
CarbonCast stands out with a strong positive slope ($127.64$, $r=0.94$), implying that pretraining can help in highly structured energy data.  
By contrast, Fitbit again shows a large negative slope ($-68.15$) with strong negative correlations, reinforcing it as an outlier due to irregular missingness.  
PEMS and Synthetic display slopes close to zero with weak correlations, further supporting the view that pretraining offers little advantage once model capacity is held constant.  
Overall, these results suggest that much of the forecasting ability stems from the large embedding head itself, with pretraining only adding value in select domains.

\begin{table}[h!]
\centering
\caption{Theil--Sen slope of Error Increase $\Delta$ (\%) versus $\Omega$ for Language Pretrained relative to Random Init, isolating the effect of pretraining.}  
\label{LangRand}
\begin{tabular}{lrrr}
\toprule
Domain & Slope & Spearman $\rho$ & Pearson $r$ \\
\midrule
Aggregate & 15.8 & 0.13 & 0.01 \\
CarbonCast & 127.64 & 1.00 & 0.94 \\
Fitbit     & -68.15 & -1.00 & -1.00 \\
PEMS       & 3.96   & 0.50 & 0.14 \\
Synthetic  & -0.28  & 0.00 & -0.03 \\
\bottomrule
\end{tabular}
\end{table}

\section{Instructive Exceptions}
\label{app:fitbit}
The Fitbit and PEMS domains show weaker alignment between error and $\Omega$ compared to other datasets.  
We attribute this to the sparsity of the two datasets. In wearable data, users often remove devices for extended periods~(eg. during sleep), creating irregular gaps. In traffic data, there are stretches of time, especially at night, where no vehicles arrive at a given intersection sensor. 
Because $\Omega$ is a purely spectral measure, it does not fully capture these missingness patterns, which may dominate the forecasting difficulty for some of the tasks chosen.  
A more thorough analysis, such as filtering for active periods, applying imputation strategies, or testing alternative error metrics, is an interesting direction for future work.  

\end{document}